\documentclass{article}

\usepackage{PRIMEarxiv}

\usepackage[utf8]{inputenc} 
\usepackage[T1]{fontenc}    
\usepackage{hyperref}       
\usepackage{url}            
\usepackage{booktabs}       
\usepackage{amsfonts}       
\usepackage{nicefrac}       
\usepackage{microtype}      
\usepackage{lipsum}
\usepackage{fancyhdr}       
\usepackage{graphicx}       
\graphicspath{{media/}}     
\usepackage{cite}
\usepackage{amsmath,amssymb,amsfonts}
\usepackage{algorithmic}
\usepackage{graphicx}
\usepackage{textcomp}
\usepackage{makecell} 
\usepackage{booktabs} 
\usepackage{array} 
\usepackage{xcolor}
\usepackage{bm}
\usepackage{multirow}
\usepackage{siunitx} 
\title{Federated Latent Factor Learning for Recovering Wireless Sensor Networks Signal with Privacy-Preserving
}
\author{
  Chengjun Yu \\
  College of Computer and Information Science \\
  Southwest University \\
  Chongqing, China \\
  \texttt{xirimuli@email.swu.edu.cn} \\
   \And
  Yixin Ran\\
  School of Engineering \\
  University of Western Australia \\
  Australia \\
  \texttt{ranyixin920@gmail.com} \\
     \And
  Yangyi Xia, Jia Wu\\
  College of Food Science \\
  Southwest University \\
  Chongqing, China \\
  \texttt{2658355128@qq.com, 2813324899@qq.com} \\
     \And
  Xiaojing Liu\\
  Chongqing True Flavor Food Co LTD \\
  Chongqing, China\\
  \texttt{845004549@qq.com} \\
}

\begin{document}
\maketitle
\begin{abstract}

Wireless Sensor Networks (WSNs) are a cutting-edge domain in the field of intelligent sensing. Due to sensor failures and energy-saving strategies, the collected data often have massive missing data, hindering subsequent analysis and decision-making. Although Latent Factor Learning (LFL) has been proven effective in recovering missing data, it fails to sufficiently consider data privacy protection. To address this issue, this paper innovatively proposes a federated latent factor learning (FLFL) based spatial signal recovery (SSR) model, named FLFL-SSR. Its main idea is two-fold: 1) it designs a sensor-level federated learning framework, where each sensor uploads only gradient updates instead of raw data to optimize the global model, and 2) it proposes a local spatial sharing strategy, allowing sensors within the same spatial region to share their latent feature vectors, capturing spatial correlations and enhancing recovery accuracy. Experimental results on two real-world WSNs datasets demonstrate that the proposed model outperforms existing federated methods in terms of recovery performance.
\end{abstract}

\keywords{Wireless Sensor Networks\and Latent Feature Analysis\and Federated Learning\and Spatial Correlation\and Low-Rank Matrix Approximation\and Data Recovery}

\section{Introduction}

Wireless Sensor Networks (WSNs) are task-oriented networks composed of multiple sensor nodes that process monitored data through embedded computing modules and transmit it to remote users or sites via wireless networks\cite{yu2022robust}. In the era of big data, WSNs are widely applied in various fields such as smart cities, industrial production, energy, network security\cite{lv2021ai}, bridge engineering\cite{putra2020multiagent}, transportation, and computer vision\cite{wang2020model}, emerging as a frontier technology in intelligent sensing. However, WSNs data are often incomplete due to sensor failures, energy-saving strategies, human errors, and other factors. As WSNs typically collect data at fixed locations and predetermined time intervals over long periods, the incomplete data\cite{b57,b59,b60} exhibit spatial and temporal correlations. These data can be modeled as a low-rank matrix, where rows represent sensor nodes, columns denote time intervals, and elements correspond to recorded WSNs data\cite{48}. Consequently, low-rank matrix approximation (LRMA)\cite{wu2024outlier,47,51,b61,b62} has been widely adopted to recover missing data in WSNs.

With the growing concern over privacy issues \cite{li2021survey}, the European Union's General Data Protection Regulation (GDPR) \cite{yang2019federated} explicitly prohibits collecting, processing, or sharing user data without consent. To ensure data security and manage resources, data owners are often unwilling to share raw data directly. This makes traditional centralized modeling methods, such as Latent Factor Analysis (LFA)\cite{17,22,23,24,43,b5,b7,b21}, less suitable for WSNs. 

Federated Learning (FL) has become an effective approach for training shared models on distributed data. It integrates information from multiple sources while protecting data privacy \cite{zhang2021survey,37}. FL has been widely used in fields like recommendation systems\cite{61} and so on \cite{Fedrec}. Recently, federated frameworks based on Matrix Factorization (MF)\cite{39,55,53} have attracted increasing attention. For example, Chai et al. proposed FedMF for secure matrix factorization \cite{FedMF}. Liang et al. further improved FedRec\cite{Fedrec} by assigning denoising clients to reduce noise in a privacy-aware manner \cite{liang2021fedrec++}. Lin et al. also incorporated meta-learning into MF to develop a federated recommendation system \cite{MetaMf}. However, current federated models have limited applications in WSNs data recovery. They often overlook the spatial characteristic of WSNs, leading to less effective data recovery. 

To address this challenge, this paper proposes a novel Federated Latent Factor Learning for Privacy-Preserving Spatial Signal Recovery model, named FLFL-SSR, aimed at recovering missing data from partially observed and privacy-preserved data in WSNs. The main ideas of FLFL-SSR are two-fold: 1) A sensor-level federated learning framework based on latent factor learning (LFL) is designed, where each sensor transmits only gradient information rather than raw data, ensuring privacy protection during model training, and 2) A local spatial sharing strategy is introduced, allowing sensors within the same spatial region to share their latent feature vectors, thereby capturing spatial correlations and enhancing recovery accuracy. With these designs, FLFL-SSR achieves both high recovery accuracy and privacy preservation.The main contributions of this work are summarized as follows:
\begin{enumerate}
    \item An FLFL-SSR model is proposed to accurately recover missing data from partially observed and privacy-preserving WSNs data.
    \item A local spatial sharing strategy is developed to capture spatial correlations, further improving the recovery performance of FLFL-SSR model.
    \item Theoretical analyses and algorithmic designs are provided to support the proposed FLFL-SSR model.
\end{enumerate}
Moreover, extensive experiments on two real-world WSNs datasets are performed to evaluate the proposed FLFL-SSR model, including comparison with state-of-the-art models and analyses of its characteristics. The results verify that FLFL-SSR has significantly higher recovery accuracy than the comparison models.

\section{Preliminaries}
\subsection{Problem of Data Recovery in WSNs}
\textbf{\emph{Definition }1 \emph{(Problem of Data Recovery in WSNs).}} Given a matrix $Y$ from WSNs consisting of $M$ rows and $N$ columns, intended for recording the data generated by $M$ sensors during $N$ time slots. Each element $y_{i,j}$ of $Y$ contains a value corresponding to the data recorded by the $i$-th sensor at the $j$-th time slot. Let $Y_K$ and $Y_U$ denote the known and unknown entry set of $Y$, respectively. The problem of data recovery in WSNs lies in estimating $Y_U$ solely based on the observations $Y_K$\cite{b58}. The value of $Y_K/Y$ is denoted as the sampling rate.
\subsection{LFA-Based Data Recovery in WSNS }
\textbf{\emph{Definition }2 \emph{(LFA-Based Data Recovery in WSNs).}} To recover \( Y_U \), the LFA model aims to learn two latent feature matrices \( P \in \mathbb{R}^{M \times k} \) and \( Q \in \mathbb{R}^{N \times k} \), where \( k \ll \min\{M, N\} \)\cite{b66,b67,b68,b69,b70,b73}. The goal is to obtain a rank-\( k \) approximation of \( Y \) by minimizing the discrepancy between the observed entries in \( Y_K \) and their estimates\cite{yuan2020multilayered,luo2020adaptive,b3,b23}. The matrix \( Y \) is approximated as:
\begin{equation}\label{1}
Y \approx PQ^T
\end{equation}
This formulation enables the model to capture the underlying structure of \( Y \) using low-dimensional representations of its rows and columns.

The \( L_2 \) norm is widely used in existing LFA models\cite{20,25,b31,b35,b72}. To avoid overfitting on \( Y_K \), Tikhonov regularization can be integrated into the objective function \( \varepsilon(P, Q) \). The objective function can be formulated as:
\begin{equation}\label{2}
\varepsilon \left( {P,Q} \right) = \left\| {J \circ \left( {Y - P{Q^{\rm T}}} \right)} \right\|_{{L_2}}^2 + \lambda \left( {\left\| P \right\|_F^2 + \left\| Q \right\|_F^2} \right)
\end{equation}
where $\circ$ denotes the Hadamard product, \(J\) is the sampling matrices, \( \| \cdot \|_{L_2} \)  denotes the $L_2$norm of a matrix\cite{21,56,58,b55,b56}, and $||\cdot||_F$ denotes the Frobenius norm of a matrix\cite{59,60,b28}.

\subsection{Federated Learning}
\textbf{\emph{Definition} 3 \emph{(Federated Learning).}} Consider an FL system comprising \( M \) sensors, where each sensor \( i \) possesses a local training dataset \( S_i \) \((i = 1, 2, \dots, M)\). Let \( S \) denote the joint training dataset, representing the union of all local datasets. The goal of FL is to collaboratively train a shared machine learning model based on \( S \) without directly sharing raw data.  

FL can be classified into three categories according to data distribution across the feature and sample spaces: horizontal FL, vertical FL, and federated transfer learning [49]. In the context of WSNs discussed in this paper, where all sensors share the same feature space but collect different samples, the FL approach falls under vertical FL.The FL training process typically consists of the following four steps:
\begin{itemize}
 \item Step 1 : The central server initializes the global model and distributes it to all sensors.  
 \item Step 2 : Each sensor trains a local model using its private data and the global model received. After training is complete, instead of sharing raw data, only the computed gradients are sent back to the server.  
 \item Step 3 : The server aggregates the gradients from all sensors to update the global model and redistributes the updated model to each sensor.  
 \item Step 4 : Steps 2 and 3 are iteratively repeated until the model converges. 
\end{itemize}

\section{THE PROPOSED FLFL-SSR MODEL}

\subsection{The Framework of FLFL-SSR}\label{}
The FLFL-SSR model is a collaborative LFA-based training framework designed for WSNs.Unlike centralized learning models, FLFL-SSR avoids requiring sensors to share raw data, thus protecting data privacy. Furthermore, by leveraging the spatial correlations intrinsic to sensor data, the FLFL-SSR model integrates the relationships into its training process, enhancing its effectiveness in data recovery.


\begin{figure}[htbp]
\centerline{\includegraphics[width=0.6\linewidth]{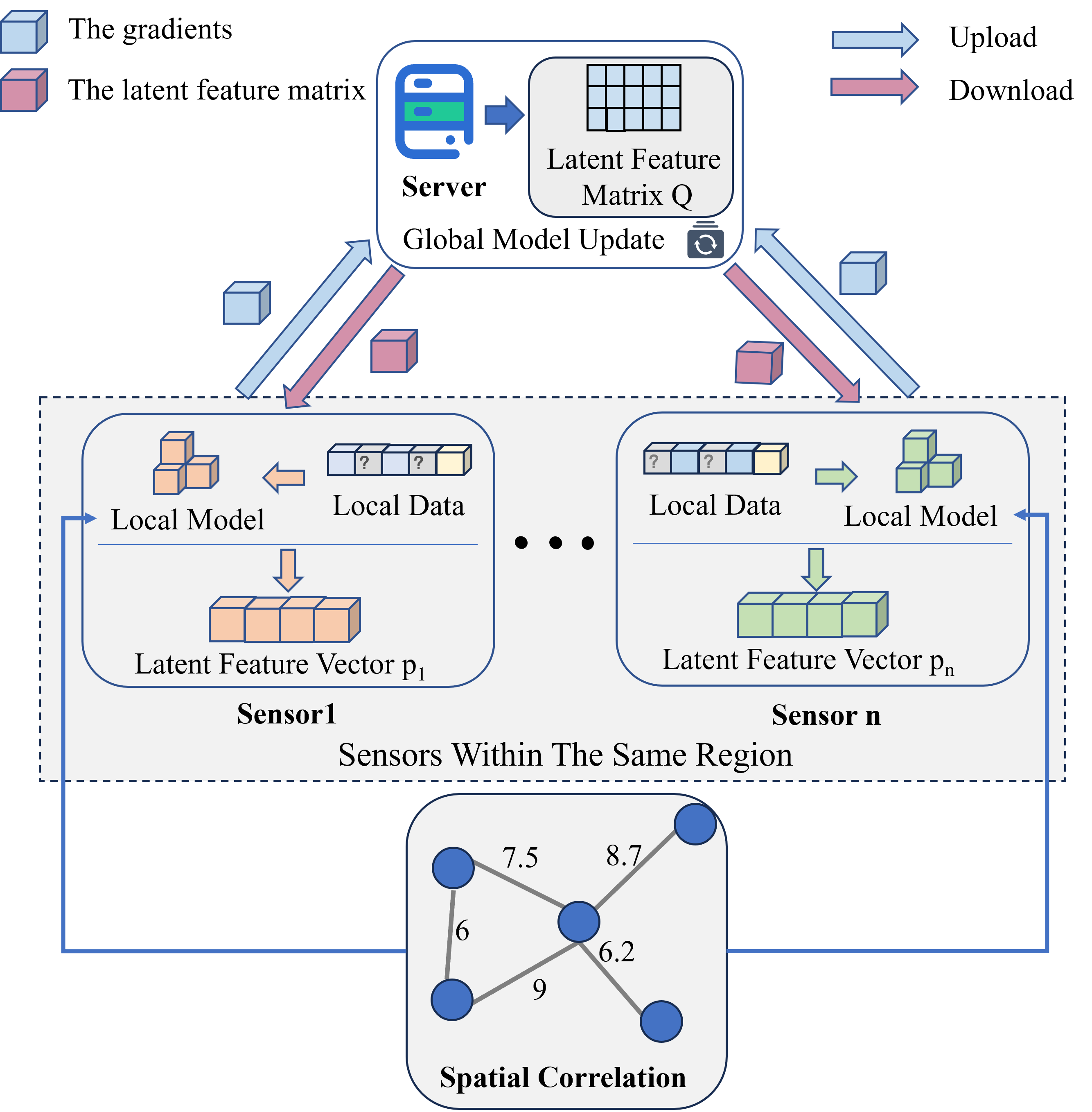}}
\caption{The Framework of FLFL-SSR}
\label{fig}
\end{figure}
 
Fig. 1 illustrates the overall architecture of the proposed FLFL-SSR model. For each sensor \( i \), the latent feature vector \( P_i \) and the required raw data for training are stored locally. Notably, sensors within the same region can share their latent feature vectors. The server, in turn, maintains only the latent feature matrix \( Q \). The coordination between the server and sensors unfolds through the following five steps:  
\begin{itemize}
\item The server randomly initializes the latent feature matrix $Q$, while each sensor randomly initializes its latent feature vector.  
\item Each sensor downloads the latent feature matrix $Q$ from the server.  
\item Each sensor performs local training using its own data, the downloaded latent feature matrix $Q$, its local latent feature vector, and the latent feature vectors shared by neighboring sensors. 
\item Sensors update their latent feature vectors and send the computed gradients to the server.  
\item The server refines the latent feature matrix $Q$ using the gradients received from all sensors.  
\end{itemize}
\subsection{Spatial Correlation in FLFL-SSR}
To address the characteristics of sensor data and ensure privacy protection, we partition the space into \( H \) regions. To capture spatial smoothness, we construct an undirected weighted sensor connectivity graph for each region, denoted as \( G_h = (V_h, E_h, W_h) \), where \( h = 1, 2, \dots, H \)\cite{48}. In this graph, \( V_h \) represents the set of vertices corresponding to the sensors within the \( h \)-th region. The number of sensors in this region is \( C_h \), meaning \( |V_h| = C_h \). The total number of sensors across all regions is \( M \), satisfying \( M = \sum_{h=1}^{H} C_h \). Additionally, \( E_h \) denotes the set of edges in the graph \( G_h \), where each edge represents the relationship between a pair of sensors.  

The weighted adjacency matrix \( W_h \) quantifies the connections between sensors, with each entry \( w_{i,i'} \) indicating the strength of the relationship between the \( i \)-th and \( i' \)-th sensors. Based on \( V_h \), we compute the distance matrix \( F_h \) using the sensors’ coordinates, where each element \( f_{i,i'}^h \) captures the Euclidean distance between the \( i \)-th and \( i' \)-th vertices. Leveraging \( F_h \), \( W_h \) is constructed by selecting the top-\( k \) nearest neighbors for each sensor, as follows:
\begin{equation}\label{3}
w_{_{i,i'}}^h = \left\{ \begin{array}{l}
\frac{1}{{{{\left( {f_{i,i'}^h} \right)}^2}}},{\rm{ }}\ if{ \rm{ }}\;  i'\in To{p_k}(i)\\
0,{\rm{     }}\qquad otherwise
\end{array} \right.,
\end{equation}
where \( \text{Top}_k(i) \) represents the set of the top-\( k \) nearest vertices to the \( i \)-th vertex. Furthermore, let \( L_h \) denote the Laplacian matrix of the graph \( G_h \). The matrix \( L_h \) is defined as follows:

\begin{equation}\label{4}
L^h = 
\scalebox{0.94}{$\displaystyle 
\begin{bmatrix}
sum(w_{1,\cdot}^h) & -w_{1,2}^h        & \cdots & -w_{1,C_h}^h \\
-w_{2,1}^h                  & sum(w_{2,\cdot}^h) & \cdots & -w_{2,C_h}^h \\
\vdots                      & \vdots            & \ddots & \vdots \\
-w_{C_h,1}^h               & -w_{C_h,2}^h      & \cdots & sum(w_{C_h,\cdot}^h)
\end{bmatrix}_{C_h \times C_h}
$} 
\end{equation}
where \( sum(w^h_{i,.}) \), where \( i \in \{1, 2, \dots, C_h\} \), represents the sum of the \( i \)-th row of the weight matrix \( W_h \). To ensure the spatial smoothness of the reconstructed matrix \( G_h \), a regularization constraint is introduced by minimizing \( \|L_h P Q^T\|_F^2 \).
\subsection{Federated Optimization}
Owing to the decentralized nature of federated learning, each sensor leverages its local data, the latent feature vectors of neighboring sensors within the same region, and the latent feature matrix provided by the server during training. Consequently, the objective function for each sensor \( i \) in the \( h \)-th region can be formulated as follows:
\begin{equation}\label{5}
\scalebox{0.93}{$\displaystyle 
\begin{split}
\varepsilon \big( p_{i,\cdot}^h,Q \big) = 
& \sum_{\substack{y_{i,j} \in Y_K^i}} \Big( y_{i,j} - p_{i,\cdot}^h q_{j,\cdot} \Big)^2 + \lambda \sum_{\substack{y_{i,j} \in Y_K^i}} \Big( \big( p_{i,\cdot}^h \big)^2 + \big( q_{j,\cdot} \big)^2 \Big) \\
& + z \sum_{\substack{y_{i,j} \in Y_K^i}} \Big( \big( L^h P^h Q^{\mathsf{T}} \big)_{(i,j)} \Big)^2
\end{split}
$} 
\end{equation}
where \( Y_i^K \) represents the set of known entries in \( Y \) for sensor \( i \), and \( P_h \) comprises all latent feature vectors of sensors within region \( h \). It is important to note that the Laplacian matrix \( L_h \) varies across different regions. Consequently, the instantaneous loss \( \varepsilon_{h}^{i,j} \) of \( \varepsilon(P, Q) \) for the \( i \)-th sensor at the \( j \)-th time slot in region \( h \), calculated on a single entry \( y_{i,j} \in Y_i^K \), can be expressed as:
\begin{equation}\label{6}
\scalebox{0.96}{$\displaystyle 
\begin{split}
\varepsilon_{_{i,j}}^{h}={\left( {{y_{i,j}} - p_{_{i,.}}^h{q_{j,.}}} \right)^2}+\lambda\left({{{\left( {p_{_{i,.}}^h} \right)}^2} +{{\left( {{q_{j,.}}} \right)}^2}}\right)
\\+{z}{\left({{{\left({{L^h}{P^h}{Q^{T}}}\right)}_{\left({i,j}\right)}}}\right)^2},
\end{split}
$} 
\end{equation}

Our objective is to derive the latent feature matrix \( Q \), stored on the server, and the latent feature matrix \( P \), maintained by the sensors. To accomplish this, we utilize the Stochastic Gradient Descent (SGD)\cite{li2023nonlinear,yuan2020generalized,li2021proportional,b26,b63,b65} algorithm to address the optimization problem\cite{18,57,b43,b44}.

\subsubsection{Federated learning in the sensor}
During the \( t \)-th iteration, each sensor \( i \) downloads the latent feature matrix \( Q \) from the server along with the local latent feature vectors of other sensors within the same region. Then, sensor \( i \) in region \( h \) calculates the gradient \( \nabla (p^h_{i,.})^t \) for its local latent feature vector and the gradient \( \nabla q^t \) for \( Q \). Based on  \eqref{6}, the element-wise computations for \( \nabla (p^h_{i,.})^t \) and \( \nabla q^t_{j,.} \). They can be expressed as follows:
\begin{equation}\label{7}
\scalebox{0.90}{$\displaystyle 
\left\{
\begin{aligned}
\nabla {\left( {p_{i,k'}^h} \right)^t} &=- 2q_{j,k'}^t\left( {y_{i,j}} - {{\left( {p_{i,.}^h} \right)}^t}q_{j,.}^t \right) \\
&\quad + 2{z_1}\left( {{{\left( {{L^h}{{\left( {{L^h}} \right)}^{\rm T}}} \right)}_{i,.}}{{\left( {{P^h}} \right)}^t}{{\left( {q_{j,.}^t} \right)}^{\rm T}}} \right)q_{j,k'}^t + 2\lambda {\left( {p_{i,k'}^h} \right)^t}, \\
\nabla q_{j,k'}^t &=- 2{\left( {p_{i,k'}^h} \right)^t}\left( {{y_{i,j}} - {{\left( {p_{i,.}^h} \right)}^t}q_{j,.}^t} \right) \\
&\quad + 2{z_1}\left( {{{\left( {{L^h}{{\left( {{L^h}} \right)}^{\rm T}}} \right)}_{i,.}}{{\left( {{P^h}} \right)}^t}{{\left( {q_{j,.}^t} \right)}^{\rm T}}} \right){\left( {p_{i,k'}^h} \right)^t} + 2\lambda q_{j,k'}^t.
\end{aligned}
\right.
$} 
\end{equation}
where \( k' \) represents the \( k' \)-th element within the latent feature dimension \( k \).

For sensor \( i \) in region \( h \), it maintains its own latent feature vector \( p^h_{i,.} \). After calculating the gradients \( \nabla (p^h_{i,.})^t \) and \( \nabla q^t \), the latent feature vector \( p^h_{i,.} \) can be updated element-wise as follows:
\begin{equation}\label{8}
{\left( {p_{i,k'}^h} \right)^{t + 1}} = {\left( {p_{i,k'}^h} \right)^t} - \eta \nabla {\left( {p_{i,k'}^h} \right)^t},
\end{equation}
where \( \eta \) denotes the learning rate of SGD, which controls the step size for each iteration of the gradient descent. After updating its local latent feature vector \( p^h_{i,.} \), sensor \( i \) transmits the gradient \( \nabla q^t \) to the server to facilitate the update of the latent feature matrix \( Q \).
\subsubsection{Federated learning in the server}
Upon receiving the gradient \( \nabla q^t \) at the \( t \)-th iteration, the server promptly updates the latent feature matrix \( Q \) using the following rule:
\begin{equation}\label{9}
    {Q^{t + 1}} = {Q^t} - \eta \nabla {q^t},
\end{equation}
So, at element-wise update of Q can be expressed as:
\begin{equation}\label{10}
    {({q_{j,k'}})^{t + 1}} = {({q_{j,k'}})^t} - \eta \nabla {({q_{j,k'}})^t},
\end{equation}
where \( (q_{j,k'})^{t+1} \) represents the \( k' \)-th element of the \( j \)-th latent feature vector in the \( t+1 \)-th iteration round.

\section{Experiments}
\subsection{General Settings}
\subsubsection{Datasets}
Two real-world benchmark datasets are selected for the experiments, both collected from WSNs — one capturing PM2.5 concentrations and the other measuring sea surface temperature. The key characteristics of these datasets are summarized in Table \ref{tab:dataset}.
\begin{table}[!ht]
    \centering
    \caption{THE PROPERTIES OF EXPERIMENTAL DATASETS} 
    \label{tab:dataset} 
    \begin{tabular}{cccccc}
        \toprule
        \thead{ \textbf{No.}} & \thead{ \textbf{Name}} & \thead{$\bm{|M|}$} & \thead{$\bm{|N|}$} & \thead{\textbf{Time}} & \thead{\textbf{Minimum/} \\ \textbf{Maximum}} \\ 
        \midrule
        D1 & \makecell{Beijing PM2.5 \\ Concentration} & 35 & 8647 & \makecell{2014-05‒\\2015-04} & \makecell{3.0/\\773.7} \\ 
        D2 & \makecell{Sea surface \\ Temperature} & 70 & 1733 & \makecell{1870-01‒\\2014-12} & \makecell{0.01/\\30.31} \\ \bottomrule
    \end{tabular}
\end{table}
\subsubsection{Evaluation Metrics}
Common evaluation metrics for assessing the accuracy of predictive models or algorithms include MAE\cite{27,b53,52,b48,b71} and RMSE\cite{26,47,b45,b46,b74}. This paper focuses on recovering missing data collected from WSNs, where the presence of outliers is inevitable. Therefore, RMSE is particularly effective in reflecting recovery accuracy in the presence of a large number of outliers. The calculation formula is as follows:
\begin{equation*}
    RMSE = \sqrt {{{\left( {\sum\limits_{{y_{i,j}} \in \gamma } {{{\left( {{y_{i,j}} - {{\hat y}_{i,j}}} \right)}^2}} } \right)} \mathord{\left/
 {\vphantom {{\left( {\sum\limits_{{y_{i,j}} \in \gamma } {{{\left( {{y_{i,j}} - {{\hat y}_{i,j}}} \right)}^2}} } \right)} \gamma }} \right.
 \kern-\nulldelimiterspace} \gamma }}
\end{equation*}
where \( \hat{y}_{i,j} \) represents the estimation of \( y_{i,j} \), and \( \gamma \) denotes the testing set. A lower RMSE value indicates higher accuracy in the model's recovery.
\subsubsection{Baselines}
We compare the proposed FLFL-SSR model with four state-of-the-art models. Table \ref{tab:baselines} gives brief descriptions of these competitors. 

\begin{table}[!ht]
    \centering
    \caption{SUMMARY OF COMPARED MODELS.} 
    \label{tab:baselines} 
    \small 
    \begin{tabular}{m{1cm}<{\centering}m{7cm}}
        \toprule
        \textbf{\makecell{Model}} & \textbf{\makecell{Description}} \\ 
        \midrule
        \makecell{FeSoG\\ \cite{Fsocgnn}}  & This method employs optimal transport distances to impute missing data values, functioning as an end-to-end machine learning approach. \\ 
        \addlinespace
        \makecell{FedMF\\ \cite{FedMF}} & This model is a federated matrix factorization approach integrated with homomorphic encryption, ensuring secure computations on encrypted data without compromising privacy. \\ 
        \addlinespace
        \makecell{FedRec++\\ \cite{liang2021fedrec++}} & This method is a novel lossless federated recommendation approach that enhances privacy by assigning denoising clients to eliminate noise in a privacy-aware manner.\\ 
        \addlinespace
        \makecell{MetaMF\\ \cite{MetaMf}}& This approach introduces a neural architecture for federated collaborative ranking, utilizing memory-based attention to effectively capture user preferences while preserving data privacy. \\ 
        \bottomrule
    \end{tabular}
\end{table}
\begin{table*}[!ht]
    \centering
    \caption{Comparison Results of Recovery Accuracy on D1 and D2 with Different Sampling Rates} 
    \label{tab:test}
    \small 
    \begin{tabular}{ccccccc}
        \toprule
        \textbf{Sampling rate} & \textbf{Metric} & \textbf{FeSoG} & \textbf{FedMF} & \textbf{FedRec++} & \textbf{MetaMF} & \textbf{FLFL-SSR} \\ \midrule
        
        \multirow{2}*{0.1} & D1 & \(\text{46.6026}_{\pm\text{0.6922}}\) & \(\text{48.0779}_{\pm\text{0.0327}}\) & \(\text{45.2432}_{\pm\text{0.0279}}\) & \(\textbf{43.1260}_{\pm\text{0.0643}}\) & \(\text{43.6354}_{\pm\text{0.1083}}\)  \\ 
        
        ~ & D2 & \(\text{2.1324}_{\pm\text{0.0163}}\) & \(\text{1.6227}_{\pm\text{0.0200}}\) & \(\text{1.5942}_{\pm\text{0.0189}}\) & \(\textbf{0.5909}_{\pm\text{0.0098}}\) & \(\text{1.1178}_{\pm\text{0.0100}}\)  \\ 
        \addlinespace
        \multirow{2}*{0.3} & D1 & \(\text{36.3763}_{\pm\text{0.8100}}\) & \(\text{32.8940}_{\pm\text{0.0527}}\) & \(\text{33.6469}_{\pm\text{0.0611}}\) & \(\text{34.7971}_{\pm\text{0.0639}}\) & \(\textbf{31.6863}_{\pm\text{0.0134}}\)  \\ 
        ~ & D2 & \(\text{1.0653}_{\pm\text{0.0081}}\) & \(\text{0.3643}_{\pm\text{0.0066}}\) & \(\text{0.3423}_{\pm\text{0.0063}}\) & \(\text{0.3796}_{\pm\text{0.0050}}\) & \(\textbf{0.2771}_{\pm\text{0.0020}}\)  \\ \addlinespace
        \multirow{2}*{0.5}& D1 & \(\text{34.4265}_{\pm\text{0.6490}}\) & \(\text{30.8070}_{\pm\text{0.0319}}\) & \(\text{29.8958}_{\pm\text{0.0162}}\) & \(\text{32.5567}_{\pm\text{0.0989}}\) & \(\textbf{29.2795}_{\pm\text{0.0209}}\)  \\ 
        ~ & D2 & \(\text{0.8522}_{\pm\text{0.0073}}\) & \(\text{0.2532}_{\pm\text{0.0011}}\) & \(\text{0.2602}_{\pm\text{0.0015}}\) & \(\text{0.3411}_{\pm\text{0.0060}}\) & \(\textbf{0.2345}_{\pm\text{0.0015}}\)  \\ \addlinespace
        \multirow{2}*{0.7} & D1 & \(\text{34.7390}_{\pm\text{0.5507}}\) & \(\text{27.7124}_{\pm\text{0.0199}}\) & \(\text{27.6353}_{\pm\text{0.0197}}\) & \(\text{31.0249}_{\pm\text{0.0522}}\) & \(\textbf{27.5861}_{\pm\text{0.0156}}\)  \\ 
        ~ & D2 & \(\text{0.7952}_{\pm\text{0.0063}}\) & \(\text{0.2204}_{\pm\text{0.0013}}\) & \(\text{0.2025}_{\pm\text{0.0007}}\) & \(\text{0.2817}_{\pm\text{0.0071}}\) & \(\textbf{0.1850}_{\pm\text{0.0006}}\)  \\ \addlinespace
        \multirow{2}*{0.9} & D1 & \(\text{32.6752}_{\pm\text{0.5710}}\) & \(\text{26.8610}_{\pm\text{0.0152}}\) & \(\text{26.6093}_{\pm\text{0.0120}}\) & \(\text{28.3891}_{\pm\text{0.1137}}\) & \(\textbf{26.4088}_{\pm\text{0.0009}}\)  \\ 
        ~ & D2 & \(\text{0.7382}_{\pm\text{0.0031}}\) & \(\text{0.1730}_{\pm\text{0.0012}}\) & \(\text{0.1690}_{\pm\text{0.0006}}\) & \(\text{0.2496}_{\pm\text{0.0040}}\) & \(\textbf{0.1618}_{\pm\text{0.0006}}\)  \\ \addlinespace\hline
        \multirow{3}*{Statistical Analysis} & win/loss & 10/0 & 10/0 & 10/0 & 8/2 & Total 38/2  \\ 
        ~ & p-value& \(\textbf{9.7656*10}^{-4}\) & \(\textbf{9.7656*10}^{-4}\) & \(\textbf{9.7656*10}^{-4}\) & \textbf{0.0527} &   \\ 
        ~ & F-rank & \textbf{4.1} & \textbf{3.1} & \textbf{2.4} & \textbf{3.4} & \textbf{1.2}  \\ 
        \bottomrule
    \end{tabular}
\end{table*}
\subsection{Performance Comparison}
This set of experiments gradually increases the sampling rate from 0.1 to 0.9 to compare the performance of all involved models. The comparison results across all datasets are shown in Table \ref{tab:test}. To further interpret these results, statistical analyses including win/loss counts, the Wilcoxon signed-ranks test \cite{45,b51}, and the Friedman test \cite{38} are performed. The win/loss analysis counts the number of cases where FLFL-SSR achieves lower or higher RMSE compared to each baseline model as the sampling rate changes. The Wilcoxon signed-ranks test, a nonparametric pairwise comparison method, assesses whether FLFL-SSR significantly outperforms each comparison model by examining the p-value's significance level. Meanwhile, the Friedman test compares the performance of multiple models across various scenarios simultaneously, using the F-rank value. A lower F-rank indicating superior recovery accuracy.

Tables \ref{tab:test} indicate that as the sampling rate increases, the RMSE values follow a decreasing trend. Additionally, several key observations can be drawn:  
\begin{itemize}
    \item When compared with federated models, FLFL-SSR shows inferior performance against MetaMF on D1 and D2, with only two cases at a sampling rate of 0.1. However, across all cases, FLFL-SSR consistently achieves the lowest RMSE among federated models.
    \item The p-values for both datasets are below 0.1, and FLFL-SSR attains the lowest F-rank value among all models, confirming that it significantly outperforms other federated models in terms of recovery accuracy.
\end{itemize}

The experimental results show that FLFL-SSR achieves the highest recovery accuracy among federated models. This is because it captures the spatial correlations in WSNs. In contrast, other federated models do not consider these correlations, which are essential for improving data recovery accuracy.

\section{CONCLUSION}
This paper proposes a federated latent factor learning (FLFL) model in spatial signal recovery for privacy-preserving Wireless Sensor Networks (WSNs), named FLFL-SSR. Our FLFL-SSR model possesses the merits of both privacy-preserving and high accuracy during data recovery in WSNs. In the experiments, FLFL-SSR is evaluated on two real world WSNs datasets. The results demonstrate that FLFL-SSR significantly outperforms the related state-of-the-art models in terms of both recovery accuracy and privacy-preserving. In future work, we plan to optimize FLFL-SSR to address the issue of outlier data present in the data collection process of WSNs while enhancing the model's robustness.

\bibliographystyle{unsrt}  
\bibliography{references}

\end{document}